\documentclass[runningheads]{llncs}
\usepackage{graphicx}
\usepackage[acronym]{glossaries}
\newacronym[plural=WSIs, firstplural=whole-slide images (WSIs)]{wsi}{WSI}{whole-slide image}
\newacronym[plural=ccRCCs, firstplural=clear cell renal cell carcinomas (ccRCCs)]{ccrcc}{ccRCC}{clear cell renal cell carcinoma}
\newacronym[plural=MLPs, firstplural=multilayer perceptrons (MLPs)]{mlp}{MLP}{multilayer perceptron}
\newacronym{cindex}{C-index}{concordance index}
\newacronym{auroc}{AUROC}{area under the receiver operating characteristic curve}
\newacronym{dfs}{DFS}{disease-free survival}
\newacronym{tcga}{TCGA}{The Cancer Genome Atlas}
\newacronym{kirc}{KIRC}{Kidney Renal Clear Cell Carcinoma Collection}
\newacronym{tcgakirc}{TCGA-KIRC}{The Cancer Genome Atlas Kidney Renal Clear Cell Carcinoma Collection}
\newacronym{ct}{CT}{computed tomography}
\newacronym{ce}{CE}{contrast-enhanced}
\usepackage{enumitem}
\usepackage{xcolor}
\usepackage{amsfonts}
\usepackage{multirow}
\usepackage{booktabs}
\usepackage{marvosym}

\begin{document}
\title{Integrating Pathology and CT Imaging for Personalized Recurrence Risk Prediction in Renal Cancer}
\titlerunning{Multimodal Prediction of Renal Cancer Recurrence}
%
\author{Daniël Boeke\inst{1,3,4}\textsuperscript{(\Letter)}\textsuperscript{*} \and Cedrik Blommestijn\inst{1,2}\textsuperscript{*} \and Rebecca N. Wray\inst{5,6,7} \and \\Kalina Chupetlovska\inst{1,4} \and Shangqi Gao \inst{5,6,7} \and Zeyu Gao\inst{5,6,7} \and \\ Regina G. H. Beets-Tan\inst{1,4} \and Mireia Crispin-Ortuzar\inst{5,6,7} \and James O. Jones\inst{5,8} \and Wilson Silva\inst{1,3} \and Ines P. Machado\inst{5,6,7}}
\authorrunning{D. Boeke et al.}

\institute{Department of Radiology, Antoni van Leeuwenhoek-Netherlands Cancer Institute, Amsterdam, The Netherlands \\ \email{d.boeke@nki.nl} \and University of Amsterdam, Amsterdam, The Netherlands \and AI Technology for Life, Department of Information and Computing Sciences, Department of Biology, Utrecht University, Utrecht, The Netherlands \and GROW Oncology, Maastricht University, Maastricht, The Netherlands \and Department of Oncology, University of Cambridge, Cambridge, UK \and Cancer Research UK Cambridge Centre, University of Cambridge, Cambridge, UK \and Early Cancer Institute, University of Cambridge, Cambridge, UK \and Cambridge University Hospitals NHS Foundation Trust, Cambridge, UK}
\maketitle              
{\renewcommand{\thefootnote}{\fnsymbol{footnote}}%
\footnotetext[1]{These authors contributed equally to this work.}}
\begin{abstract}
Recurrence risk estimation in clear cell renal cell carcinoma (ccRCC) is essential for guiding postoperative surveillance and treatment. The Leibovich score remains widely used for stratifying distant recurrence risk but offers limited patient-level resolution and excludes imaging information. This study evaluates multimodal recurrence prediction by integrating preoperative computed tomography (CT) and postoperative histopathology whole-slide images (WSIs). A modular deep learning framework with pretrained encoders and Cox-based survival modeling was tested across unimodal, late fusion, and intermediate fusion setups. In a real-world ccRCC cohort, WSI-based models consistently outperformed CT-only models, underscoring the prognostic strength of pathology. Intermediate fusion further improved performance, with the best model (TITAN--CONCH with ResNet-18) approaching the adjusted Leibovich score. Random tie-breaking narrowed the gap between the clinical baseline and learned models, suggesting discretization may overstate individualized performance. Using simple embedding concatenation, radiology added value primarily through fusion. These findings demonstrate the feasibility of foundation model-based multimodal integration for personalized ccRCC risk prediction. Future work should explore more expressive fusion strategies, larger multimodal datasets, and general-purpose CT encoders to better match pathology modeling capacity.

\end{abstract}


\section{Introduction}

\Gls{ccrcc} is the most common subtype of renal cancer and carries a substantial risk of recurrence after surgery, even when radical nephrectomy is performed with curative intent \cite{Capitanio2016}. Estimating recurrence risk is important for determining patient-specific follow-up schedules and the potential use of adjuvant therapies. In clinical practice, the Leibovich score is widely used to stratify the risk of distant recurrence based on a limited set of histopathologic features, including tumor stage, size, grade, and presence of necrosis \cite{Leibovich2003,Leibovich2018}. Although clinically validated and easy to apply, this score provides only a population-level risk estimate, relies on pathologist evaluations that can be labor-intensive, and may lack the resolution needed for personalized decision-making.

Previous studies have proposed data-driven approaches to improve risk prediction. Margue et al.~\cite{Margue2024} showed that structured clinical and pathology report data can be used to expand the predictive feature set beyond that of the Leibovich score, allowing for more individualized risk estimation. Gui et al.~\cite{Gui2023} combined clinical, genomic, and pathological \glspl{wsi} in a multimodal framework and found that this integration improved performance, with \gls{wsi}-derived features contributing most strongly to the model. These findings support the idea that integrating complementary data sources can enhance recurrence prediction. However, to our knowledge, no prior study has systematically combined \gls{wsi} and radiological imaging for this task, despite their widespread availability and potential complementarity.

Recent advances in foundation models trained on large-scale medical imaging datasets have made it increasingly feasible to integrate multiple imaging modalities. These models provide versatile feature representations that can be transferred to new tasks, enabling effective multimodal integration even when training data is limited. This study investigates the use of both postoperative \gls{wsi} and preoperative \gls{ct} imaging for distant recurrence risk prediction in \gls{ccrcc}. A modular deep learning pipeline is developed using feature representations derived from foundation models trained on large-scale medical imaging datasets. Both unimodal and multimodal settings are explored, incorporating intermediate and late fusion techniques. Model performance is benchmarked against the clinical Leibovich score to evaluate whether imaging-based integration can provide more individualized recurrence risk estimates. The approach is implemented on a publicly available dataset, enabling a systematic exploration of imaging modality combinations in a realistic clinical context. This work aims to support the development of clinically meaningful, data-driven decision support tools by integrating multimodal data to enable more personalised follow-up strategies. The main contributions of this work are as follows:

\begin{enumerate}[label=\textbf{\arabic*.}]
    \item \textbf{A modular multimodal modeling framework} that leverages foundation models to extract meaningful feature representations from \gls{wsi} and \gls{ct} data, enabling flexible experimentation with both late and intermediate fusion strategies under limited supervision.
    
    \item \textbf{Interpretability analysis of modality-specific and fused representations} for recurrence risk prediction in \gls{ccrcc}, including performance comparisons against unimodal baselines and the clinically established Leibovich score.
    
    \item \textbf{Benchmarking on a publicly available, real-world cohort} to provide a reproducible evaluation of fusion strategies and assess the complementary prognostic value of anatomical and histopathological imaging in a realistic clinical setting.
\end{enumerate}

\section{Methodology}
\begin{figure}[t!]
    \centering
    \includegraphics[width=\textwidth]{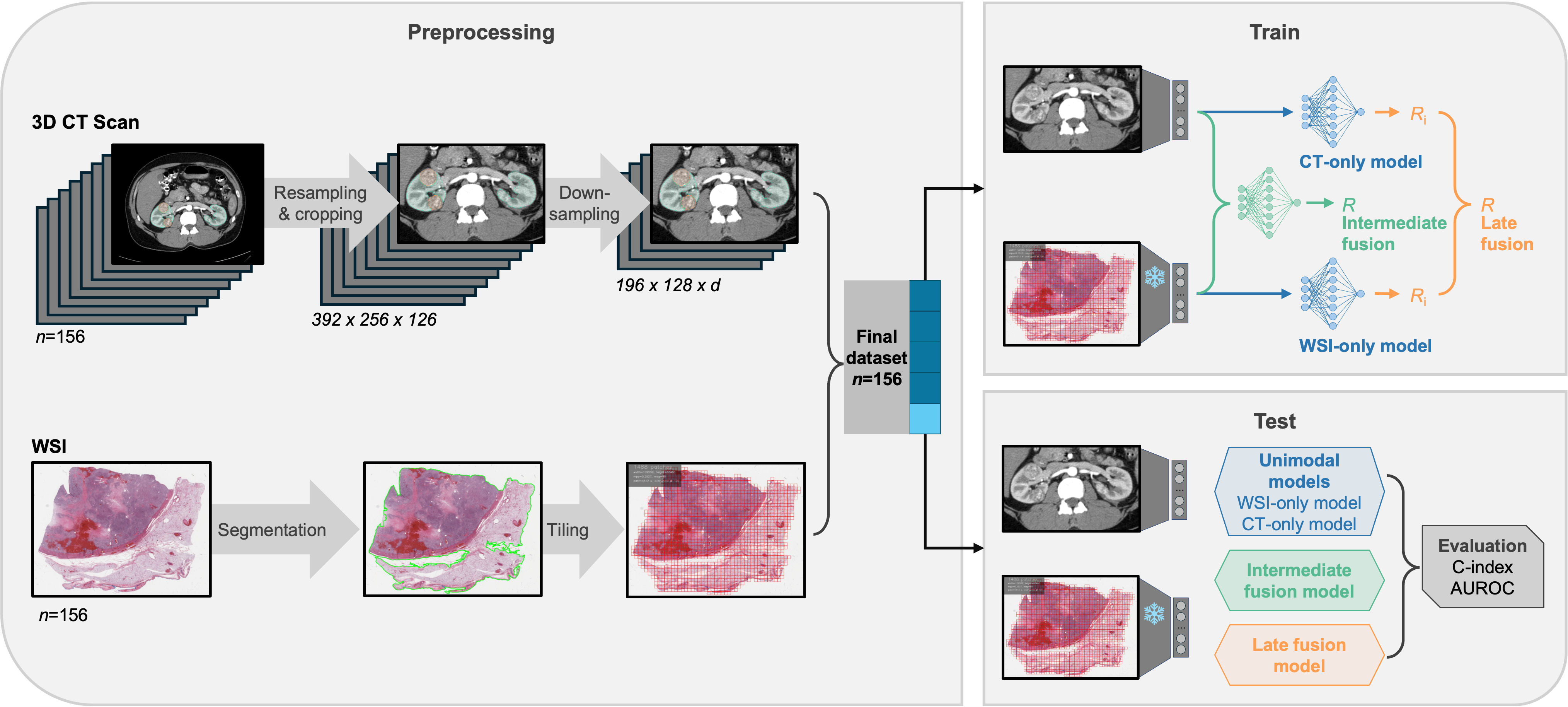}
    \caption{
    Overview of the model pipeline. Preoperative \gls{ct} and resected pathology \glspl{wsi} are independently preprocessed and encoded into patient-level feature vectors. These vectors are used in unimodal survival models or combined using intermediate fusion via embedding concatenation or late fusion strategies for recurrence risk prediction. Manual segmentations, with kidneys in green and tumors in orange, were not used during model training and are shown only for visual clarity.}

    \label{fig:workflow}
\end{figure}

Figure~\ref{fig:workflow} provides an overview of the proposed modeling pipeline of this study. The following sections describe the details of feature representation, fusion strategies, and outcome modeling.\\

\noindent\textbf{Cross-Modality Feature Representation.} 
\Gls{wsi} representations are derived using the Trident framework \cite{Zhang2025,Vaidya2025}, which performs tissue segmentation, patch tiling, and slide-level aggregation. Two encoder combinations are employed: the patch encoder CONCHv1.5 \cite{Lu2024} with the slide encoder TITAN \cite{Ding2024}, and the patch encoder CTransPath \cite{Wang2022} with the slide encoder CHIEF \cite{Wang2024}. In both cases, patch embeddings are aggregated into a single patient-level vector by the corresponding slide encoder. All pathology encoders are kept frozen during training due to their scale and pretraining complexity. 
\gls{ct} features are extracted using pretrained 3D encoders applied to resampled, cropped, and intensity-normalized volumes. Two architectures are used: MedicalNet \cite{Chen2019}, a 3D ResNet pretrained on diverse \gls{ct} and MRI datasets, and SwinUNETR \cite{Hatamizadeh2022}, a hybrid transformer with U-Net--style decoding pretrained on BraTS brain MRI data~\cite{Baid2023}. In both cases, the segmentation head is removed, and the final encoder block is used to generate a volume-level feature vector. These encoders are fine-tuned for recurrence prediction: for MedicalNet, only the final residual block is updated; for SwinUNETR, only the CNN encoder component is trained. Fine-tuning is feasible due to the smaller model size relative to pathology encoders. Each \gls{ct} scan is represented by a fixed-length vector, obtained via a fully connected layer that maps the encoding to the \gls{wsi} embedding dimension, ensuring balanced representation across modalities.\\

\noindent\textbf{Unified Prediction Framework.} All models use a Cox-based \gls{mlp} prediction architecture that maps input features to a scalar risk score. The \gls{mlp} consists of an input layer, one hidden layer with ReLU activation and layer normalization, and an output layer producing the final prediction. While input dimensionality varies by modality or fusion strategy, the overall structure stays the same. Each model variant is trained separately using the Cox partial likelihood loss, with independent tuning of the randomly initialized prediction head to fit modality-specific representations. In the unimodal setting, patient-level feature vectors from either \gls{ct} or \gls{wsi} data are passed independently through modality-specific \glspl{mlp}. These models are trained separately and serve as baselines to assess the predictive value of each imaging modality alone. In the late fusion approach, \gls{ct} and \gls{wsi} models are first trained independently. Their output risk scores \( R_{\textbf{i}}\) are then combined post hoc using a weighted average: \( R = \alpha \cdot R_{\text{WSI}} + (1 - \alpha) \cdot R_{\text{CT}} \), where \( \alpha \in [0,1] \) controls the relative contribution of each modality. In the intermediate fusion strategy, \gls{ct} and \gls{wsi} feature vectors are concatenated into a joint representation, which is passed through a single \gls{mlp}. This setup enables the model to learn feature interactions between modalities and capture complementary information during training.\\


\noindent\textbf{Survival Outcome Modeling via Cox Partial Likelihood Estimation.} To estimate recurrence risk from imaging-derived features, all models are trained using a Cox proportional hazards framework \cite{Cox1972}. This approach has also been adopted in deep learning-based survival models, such as DeepSurv \cite{Katzman2018}, which optimize the Cox partial likelihood using neural networks. Given a set of patients indexed by \( i = 1, \dots, N \), each with a feature vector \( \mathbf{x}_i \), the model outputs a scalar risk score \( f(\mathbf{x}_i) \in \mathbb{R} \). A patient's hazard function \( h(t \mid \mathbf{x}_i) \) is defined as:
\begin{equation}
h(t \mid \mathbf{x}_i) = h_0(t) \exp(f(\mathbf{x}_i))
\end{equation}
where \( h_0(t) \) is an unspecified baseline hazard function, and \( \exp(f(\mathbf{x}_i)) \) represents the relative risk associated with the patient's covariates. This formulation captures the ordering of recurrence events without requiring precise estimation of survival times. In clinical datasets, not all patients experience recurrence within the observation window, making it essential to account for censoring when learning from event times. Model training is therefore based on a Cox partial likelihood loss objective, which for uncensored recurrence times \( T_i \) is given by:
\begin{equation}
\mathcal{L}_{\text{Cox}} = - \sum_{i: \delta_i = 1} \left( f(\mathbf{x}_i) - \log \sum_{j \in \mathcal{R}(T_i)} \exp(f(\mathbf{x}_j)) \right),
\end{equation}
where \( \delta_i \in \{0, 1\} \) is the event indicator, and \( \mathcal{R}(T_i) \) denotes the risk set, which includes all patients still at risk at time \( T_i \). This loss function encourages the model to assign higher risk scores to patients who experience earlier recurrence, thereby preserving a concordant ordering with observed outcomes.\\


\section{Experiments}
\noindent\textbf{Datasets.} \gls{tcgakirc} dataset~\cite{TheCancerGenomeAtlasResearchNetwork2013,Akin2016} includes \gls{ce}-\gls{ct} scans, resected \glspl{wsi}, and clinical annotations with postoperative distant recurrence outcomes, making it suitable for \gls{ccrcc} recurrence evaluation. From the full cohort, 187 patients with both \gls{ct} and \gls{wsi} data were identified. After manual review, 156 patients were retained based on the availability of abdominal \glspl{ct} with sufficient image quality, appropriate contrast phase (late arterial or early venous), and visible kidney structures. The final cohort had complete imaging and survival data suitable for time-to-event modeling. Patients had a median age of 57.6 years (range 26–81), with 70.5\% male. Recurrence occurred in 40 patients (25.6\%), while 116 (74.4\%) remained disease-free during follow-up. Tumor staging was mainly T1 (55.2\%) and T3 (30.1\%), with rare nodal involvement (1.9\% N1) and undetermined nodal status in 83 patients (53.2\%). Histologic grades were mostly 2 (42.3\%) and 3 (46.8\%).\\

\noindent\textbf{Experimental Protocol}. Experiments were conducted using nested stratified cross-validation with five outer folds and three inner folds. Stratification was based on recurrence status, and fold assignments were fixed across all experiments to ensure consistent evaluation across model types. To focus the model on anatomically relevant regions, 3D \gls{ct} volumes were first resampled to a uniform voxel spacing based on the average resolution across the dataset. Kidney masks were then generated using a \texttt{nnU-NetV2} model~\cite{Isensee2018} trained on the KiTS21 dataset~\cite{Heller2023}. The largest kidney volume in the training set was used to define a fixed bounding box of size \(392 \times 256 \times 126\), which was consistently applied to all resampled images by cropping around the kidneys. Volumes were subsequently downsampled to \(196 \times 128 \times 8\) for MedicalNet \cite{Chen2019} and \(196 \times 128 \times 32\) for SwinUNETR \cite{Hatamizadeh2022} to match the input requirements of the respective pretrained encoders. This process standardized spatial resolution, focused attention on the kidney region, and reduced computational cost while preserving relevant anatomy. Diagnostic-quality \glspl{wsi} from the \gls{tcgakirc} dataset were selected based on expert-rated slide annotations. Preprocessing was performed using the Trident framework with default parameters. All slides were processed at the default magnification level of 20, ensuring consistent spatial resolution across patients and compatibility with the pretrained encoders. 

Training was conducted using the Adam optimizer with a cosine annealing scheduler and linear warmup. In the inner cross-validation loop, models were trained for up to 200 epochs with early stopping based on validation \gls{cindex}, using a patience of 10 epochs. The optimal number of epochs was selected per outer fold based on inner-loop performance and used to train the final model on the corresponding outer training set. Bayesian hyperparameter optimization was performed using Optuna~\cite{Akiba2019}, targeting the average validation \gls{cindex} across inner folds. Tuned hyperparameters and their respective search ranges included: learning rate \([10^{-5}, 10^{-3}]\), weight decay \([10^{-6}, 10^{-2}]\), L1 penalty \([10^{-6}, 10^{-2}]\), hidden layer size \([32, 512]\), dropout rate \([0.0, 0.5]\), and cosine scheduler floor \([10^{-6}, 10^{-2}]\). For late fusion models, the fusion weight \(\alpha\), controlling the contribution of each modality, was also tuned over the range \([0, 1]\). All models were trained with a batch size of 16, and final evaluation was performed using the best inner-fold configuration for each outer fold.\\

\noindent\textbf{Adjusted Leibovich score as a Clinical Baseline for Risk Stratification.} The Leibovich score is a clinically validated tool for estimating the risk of non-local recurrence in \gls{ccrcc} following nephrectomy. It integrates pathological features including tumor size, nuclear grade, T-stage, lymph node involvement, and tumor necrosis~\cite{Leibovich2003,Leibovich2018}, and remains widely used in clinical practice~\cite{Blackmur2021}. In this study, the score was computed from structured pathology and surgical data. Necrosis was omitted due to its absence in the dataset, resulting in an adjusted score. Accordingly, standard risk group thresholds (i.e., low, intermediate, and high) were not applied, as they may not be valid without the necrosis component. The adjusted score ranged from 0 to 9, with 41.0\% of patients scoring below 3 and 91.0\% scoring below 6. This discrete score was used as a baseline risk estimate and evaluated using the same time-to-event metrics as the learned models, enabling consistent comparison of recurrence risk discrimination.\\

\noindent\textbf{Evaluation Metrics.} The primary evaluation metric is the \gls{cindex}~\cite{Harrell1982}, a generalization of the \gls{auroc}~\cite{Hanley1982} that accounts for right-censored data. It quantifies the model’s ability to accurately rank patients by recurrence risk. The \gls{cindex} is defined as:
\begin{equation}
    \text{C-index} = \frac{
    \sum_{i,j} 1_{\{T_j < T_i\}} \cdot 1_{\{\eta_j > \eta_i\}} \cdot \delta_j
}{
    \sum_{i,j} 1_{\{T_j < T_i\}} \cdot \delta_j
}
\end{equation}
where \( T_i \) is the observed time for patient \( i \), \( \delta_i \in \{0,1\} \) is the event indicator, and \( \eta_i = f(\mathbf{x}_i) \) is the predicted risk score. Tied predictions are assigned a concordance of 0.5, following the implementation in the Python package \texttt{lifelines}~\cite{Davidson-Pilon2019}. A \gls{cindex} of 1.0 indicates perfect ranking, while 0.5 indicates random performance. The \gls{auroc} for 5-year \gls{dfs} is reported as a secondary metric to align with prior studies. Patients who experience recurrence within 5 years are labeled positive; those who remain recurrence-free beyond 5 years, including censored patients, are labeled negative. The \gls{auroc} evaluates the model’s ability to separate these two groups:
\begin{equation}
    \text{AUROC}_{t=5} = 
\frac{
    \sum_{i, j} \; [Y_i(5) = 1 \land Y_j(5) = 0 \land \eta_i > \eta_j]
}{
    \sum_{i, j} \; [Y_i(5) = 1 \land Y_j(5) = 0]
}
\end{equation}
where \( Y_i(5) \in \{0, 1\} \) is the binary indicator of whether patient \( i \) experienced recurrence within five years.

\section{Results}
\noindent\textbf{Predictive Model Performance.} Table~\ref{tab:final-results} summarizes predictive performance across unimodal models, fusion strategies, and the clinical baseline using five-fold cross-validation and described evaluation metrics. The Leibovich score achieved the highest overall performance (\gls{cindex} \(0.8046 \pm 0.0348\); AUROC \(0.8139 \pm 0.1412\)), consistent with its established role in postoperative risk stratification. However, its discretized nature results in many patients receiving identical scores, which limits granularity and potentially inflates the \gls{cindex} under standard tie-handling. When ties were instead resolved randomly, closer to how such scores would be used in practice, the mean \gls{cindex} dropped to \(0.7494 \pm 0.0430\), more closely aligning with the top-performing learned models. This demonstrates that, despite its overall clinical effectiveness, the discretized nature of the Leibovich score may limit its utility in personalized decision support.\\

\begin{table}[t!]
  \centering
  \caption{
C-index and AUROC for unimodal models, multimodal fusion strategies, and clinical baseline. Best learned results are in \textbf{bold}, second-best are \underline{underlined}. Leibovich shown in \textbf{bold}$^\dagger$ to denote highest overall performance; Leibovich (RT) uses randomized tie-breaking to adjust for score discretization.
}

  \label{tab:final-results}
  \begin{tabular}{ccccc}
    \toprule
    \textbf{Strategy} & \textbf{WSI Encoder} & \textbf{CT Encoder} & \textbf{C-index $\uparrow$} & \textbf{AUROC $\uparrow$} \\
    \midrule

    \multirow{5}{*}{Unimodal}
      & TITAN--CONCH        & --           & 0.745±0.046 & 0.771±0.081 \\
      & CHIEF--CTransPath   & --           & 0.663±0.054 & 0.601±0.152 \\
      & --                  & ResNet-10   & 0.547±0.119 & 0.623±0.141 \\
      & --                  & ResNet-18   & 0.518±0.124 & 0.532±0.200 \\
      & --                  & SwinUNETR   & 0.549±0.129 & 0.604±0.154 \\
    \midrule

    \multirow{6}{*}{Late}
      & \multirow{3}{*}{TITAN--CONCH}
        & ResNet-10         & 0.727±0.106 & 0.763±0.088 \\
      &                     & ResNet-18         & 0.732±0.048 & 0.741±0.088 \\
      &                     & SwinUNETR         & 0.739±0.069 & 0.748±0.110 \\
      \cmidrule(lr){2-5}
      & \multirow{3}{*}{CHIEF--CTransPath}
        & ResNet-10         & 0.692±0.068 & 0.597±0.162 \\
      &                     & ResNet-18         & 0.666±0.046 & 0.627±0.190 \\
      &                     & SwinUNETR         & 0.692±0.029 & 0.606±0.153 \\
    \midrule

    \multirow{6}{*}{Intermediate}
      & \multirow{3}{*}{TITAN--CONCH}
        & ResNet-10         & 0.742±0.086 & \textbf{0.780±0.133} \\
      &                     & ResNet-18         & \textbf{0.775±0.044} & \underline{0.774±0.115} \\
      &                     & SwinUNETR         & \underline{0.749±0.083} & 0.768±0.063 \\
      \cmidrule(lr){2-5}
      & \multirow{3}{*}{CHIEF--CTransPath}
        & ResNet-10         & 0.691±0.122 & 0.671±0.093 \\
      &                     & ResNet-18         & 0.669±0.112 & 0.649±0.170 \\
      &                     & SwinUNETR         & 0.679±0.068 & 0.624±0.201 \\
    \midrule
    Leibovich & -- & -- & \textbf{0.805±0.035}$^\dagger$ & \textbf{0.814±0.141}$^\dagger$ \\
    Leibovich (RT) & -- & -- & \underline{0.749±0.043} & \textbf{0.814±0.141}$^\dagger$ \\
    \bottomrule
  \end{tabular}
\end{table}

\noindent\textbf{Performance Analysis of Unimodal and Multimodal Strategies.} Among learned unimodal models, those based on histopathology consistently outperformed radiology-based models. This is not unexpected, as many established prognostic features are histologic in nature. In contrast, \gls{ct}-derived models showed limited predictive value, possibly due to domain mismatch in pretraining, which fine-tuning may not fully overcome given limited data. Most radiology encoders were originally developed for segmentation tasks, which may limit their effectiveness for survival prediction compared to pathology foundation models, which are trained for general-purpose feature extraction across tasks. 

Fusion strategies, however, revealed that \gls{ct} can still provide complementary signal when combined with pathology. In the multimodal late fusion setting, fusion weights across models consistently assigned high importance to \gls{wsi}-derived predictions, with \(\alpha\)-values ranging from 0.8 to 0.9, indicating that pathology contributed the majority of the predictive signal. Importantly, intermediate fusion consistently outperformed both unimodal and late fusion models, with all \gls{wsi}--\gls{ct} combinations matching or exceeding their \gls{wsi}-only baselines. The best result was achieved by the combination of TITAN--CONCH and the ResNet-18 model (\gls{cindex} \(0.775 \pm 0.044\); \gls{auroc} \(0.774 \pm 0.115\)).

Models using the TITAN--CONCH pathology encoder consistently achieved the strongest performance, suggesting this combination is particularly effective at capturing histologic risk patterns relevant to recurrence. Among \gls{ct} encoders, ResNet-18 outperformed both the shallower ResNet-10 and the transformer-based SwinUNETR in intermediate fusion, despite being the lowest-performing model in the unimodal \gls{ct} setting. This discrepancy highlights substantial variability in \gls{ct} model performance and suggests that certain encoders may only reveal their predictive value when combined with complementary modalities. The relatively high standard deviations observed across \gls{ct}-based models further underscore the instability of radiologic representations in this setting.

While some intermediate fusion models approached the clinical baseline and outperformed their unimodal and late fusion counterparts, these gains should be interpreted with caution due to their reported standard deviations. All fusion models used a simple concatenation-based strategy, suggesting that more expressive techniques such as cross-attention or co-learning architectures may further enhance integration and performance. Although pretrained encoders helped mitigate data scarcity, particularly for the information-rich pathology modality, the radiology side may benefit from the development of general-purpose foundation models trained across diverse tasks and anatomical regions. This could enhance the robustness of radiologic representations and align them more closely with the modeling capabilities observed in pathology. Future work could also explore more advanced fusion mechanisms, richer multimodal datasets, and the incorporation of additional data types to further enhance personalized recurrence risk prediction.\\

\noindent\textbf{Case-based Analysis of Prediction Discrepancies.}
\begin{figure}[t!]
\centering\includegraphics[width=\textwidth]{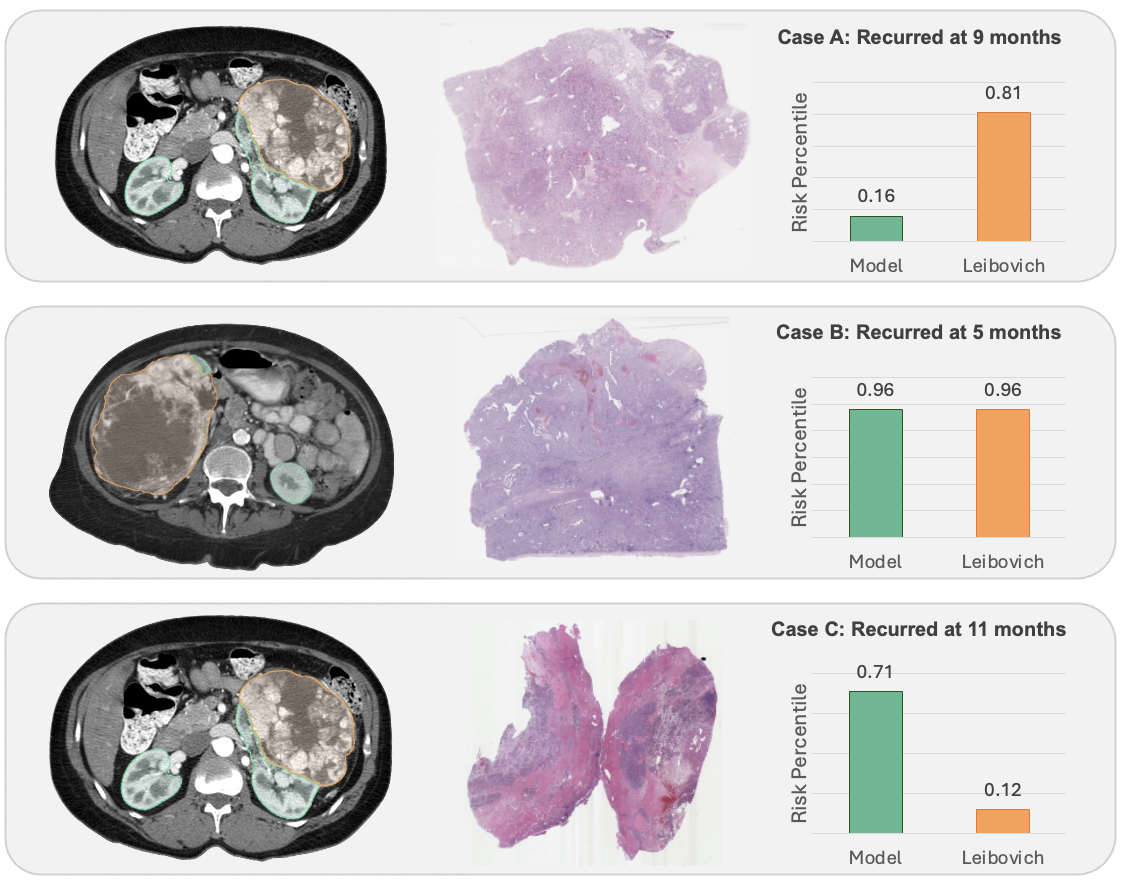}
    \caption{
    Case examples illustrating predicted recurrence risk from the best-performing model, intermediate fusion of TITAN--CONCH and ResNet-18, compared to the adjusted Leibovich score. Risk scores are shown as percentiles within the full patient cohort. All patients experienced recurrence. Bars indicate each method’s estimated relative risk. Kidney tissue is delineated in green and tumor tissue in orange on CT.}
    \label{fig:cases}
\end{figure}
A preliminary review of three representative cases was conducted to illustrate different patterns of concordance between the multimodal model and the Leibovich score: one where the clinical score performed better, one showing agreement, and one where the model outperformed the clinical score (Fig.~\ref{fig:cases}). Predictions are shown for the best-performing model: intermediate fusion of TITAN--CONCH and ResNet-18. 
In Case A, the model substantially underestimated recurrence risk, despite an event occurring within 9 months. The selected WSI lacked overt high-risk features, although the pathology report noted necrosis and high Fuhrman grade, suggesting the slide may not have captured the most aggressive tumor regions. While the CT scan showed extensive disease, the radiology encodings may not have conveyed this effectively, highlighting potential limitations in both spatial sampling and feature representation. 
Case B showed strong concordance, with both model and score predicting high risk, consistent with early recurrence. Apparent perinephric fat invasion was observed, making this a relatively straightforward case where classical pathologic and radiologic features aligned well with the clinical outcome. 
In contrast, Case C illustrates a scenario where the model predicted high risk not reflected in the Leibovich score. This discrepancy may reflect the model’s ability to capture subtle histologic or radiologic signals not explicitly encoded in structured clinical variables, underscoring the potential of learned, data-driven representations to complement existing tools.

\section{Conclusion}
This study investigated postoperative distant recurrence risk prediction in \gls{ccrcc} using a multimodal imaging framework that combines preoperative \gls{ct} and postoperative \gls{wsi} data. A modular pipeline with pretrained encoders and Cox-based survival models was evaluated in unimodal and multimodal late and intermediate fusion settings, with an adjusted Leibovich score as a benchmark. \gls{wsi}-based models consistently outperformed \gls{ct}-only models, reflecting the substantial prognostic value of pathology. Intermediate fusion further improved performance, with the best learned model, TITAN–CONCH combined with ResNet-18, approaching that of the adjusted Leibovich score. This suggests that \gls{ct} and \gls{wsi} offer complementary prognostic information that, when integrated, can enhance risk estimation. The discretized nature of the Leibovich score may overstate its performance for individualized prediction, as random tie-breaking brought its concordance closer to the learned models. These results were achieved using basic embedding concatenation, with strong fusion performance even from weaker unimodal \gls{ct} encoders such as ResNet-18. This highlights the untapped potential in radiologic data when effectively integrated. Dataset size likely limited model performance, particularly for \gls{ct}, and contributed to encoder variability. Future work should explore more expressive fusion strategies, such as cross-attention or co-learning, and develop general-purpose \gls{ct} foundation models analogous to those in pathology. Scaling to larger multimodal datasets and improving interpretability will be key for enabling clinically trustworthy, personalized decision support systems.

{\fontsize{9}{12}\selectfont
\subsubsection{Acknowledgements.} Research at the Netherlands Cancer Institute is supported by the Dutch Cancer Society and the Dutch Ministry of Health, Welfare and Sport. This publication is part of the project ”Ordinality-informed Federated Learning for Robust and Explainable Radiology AI” with file number NGF.1609.241.009 of the research programme AiNED XS Europa, which is (partly) financed by the Dutch Research Council (NWO). Additional support was provided by the Cancer Research UK Cambridge Centre [CTRQQR-2021\textbackslash100012] and The Mark Foundation for Cancer Research [RG95043].
}

{\fontsize{9}{12}\selectfont
\subsubsection{Disclosure of Interests.}
The authors have no competing interests to declare that are relevant to the content of this article.
}
\bibliographystyle{splncs04}
\bibliography{Misc/mybibliography}

\end{document}